\documentclass{article}
\usepackage{ijcai25}

\usepackage{times}
\usepackage{soul}
\usepackage{url}
\usepackage[hidelinks]{hyperref}
\usepackage[utf8]{inputenc}
\usepackage[small]{caption}
\usepackage{graphicx}
\usepackage{amsmath}
\usepackage{amsthm}
\usepackage{booktabs}
\usepackage{algorithm}
\usepackage{algorithmic}
\usepackage[switch]{lineno}

\usepackage{multirow}
\usepackage{subcaption}
\usepackage{amsmath}
\usepackage{longtable}
\newcommand{\methodname}{{\tt{UncertainXFL}}}

\title{Uncertainty-Aware Explainable Federated Learning}

\author{
Yanci Zhang\And
Han Yu
\affiliations
College of Computing and Data Science, Nanyang Technological University, Singapore
\emails
\{yanci001, han.yu\}@ntu.edu.sg
}

\begin{document}

\maketitle

\begin{abstract}

Federated Learning (FL) is a collaborative machine learning paradigm for enhancing data privacy preservation. Its privacy-preserving nature complicates the explanation of the decision-making processes and the evaluation of the reliability of the generated explanations. In this paper, we propose the \underline{Uncertain}ty-aware e\underline{X}plainable \underline{F}ederated \underline{L}earning (\methodname{}) to address these challenges. It generates explanations for decision-making processes under FL settings and provides information regarding the uncertainty of these explanations. \methodname{} is the first framework to explicitly offer uncertainty evaluation for explanations within the FL context. Explanatory information is initially generated by the FL clients and then aggregated by the server in a comprehensive and conflict-free manner during FL training. The quality of the explanations, including the uncertainty score and tested validity, guides the FL training process by prioritizing clients with the most reliable explanations through higher weights during model aggregation. Extensive experimental evaluation results demonstrate that \methodname{} achieves superior model accuracy and explanation accuracy, surpassing the current state-of-the-art model that does not incorporate uncertainty information by $2.71\%$ and $1.77\%$, respectively. By integrating and quantifying uncertainty in the data into the explanation process, \methodname{} not only clearly presents the explanation alongside its uncertainty, but also leverages this uncertainty to guide the FL training process, thereby enhancing the robustness and reliability of the resulting models.

\end{abstract}

\section{Introduction}

In recent years, federated learning (FL)~\cite{Kairouz-et-al:2021} has emerged as an important approach that enables multiple parties to collaboratively train a shared model, while preserving local data privacy. Unlike traditional machine learning (ML) methods that require data to be sent to a central server, FL involves training models locally and only transferring the model updates to the FL server. The server then aggregates these updates to enhance the global model before redistributing it back to the participants. This method not only preserves privacy but also minimizes the transfer of potentially sensitive information, making it promising for compliance with data protection laws like GDPR~\cite{GDPR}. FL is increasingly applied in areas where data privacy is critical (e.g., healthcare, finance).

For mission critical applications, it is essential that FL models are not only privacy-preserving, but also transparent to facilitate stakeholder understanding and trust building. Explainable Artificial Intelligence (XAI)~\cite{Yu-et-al:2014AAMAS,gunning2019xai,xu2019explainable} enhances this transparency by making the decision processes of AI models accessible. XAI addresses the complexities of ``black box'' ML models by providing insights into their decision processes. Common methods in XAI include visual explanations that highlight key features~\cite{selvaraju2017grad}, feature importance scores that quantify the impact of the inputs~\cite{ribeiro2016should,lundberg2017unified,koh2020concept}, and logical rules~\cite{barbiero2022entropy,zhang2024lr} that outline model reasoning. 

Existing XAI methods do not account for the uncertainty inherent in AI predictions, due to factors such as noise in the data, incomplete information and the limitations of the model itself. Incorporating uncertainty into XAI is crucial for enhancing the reliability of these systems~\cite{schum2014toward,kochenderfer2015decision,seuss2021bridging,seoni2023application}. By equipping XAI explanations with uncertainty, users can gain valuable insights into the confidence level of the decisions made by AI systems. This enhancement helps in providing a clearer understanding of the model's capabilities and limitations, facilitating more informed and cautious decision-making, especially in mission critical applications where the stakes of AI decisions are high.

Integrating XAI into FL while considering uncertainty faces significant technical challenges:
\begin{enumerate}
    \item \textbf{Expression of Uncertainty in Explanations:} Accurately quantifying and presenting uncertainty information in explanations in a way that is both informative and easy for users to understand is challenging.

    \item \textbf{Formation of Global, Conflict-Free Explanations:} The distributed nature of FL leads to diversity in data quality and completeness across data owners (a.k.a., FL clients), hindering the development of consistent and reliable explanations. The challenge lies in aggregating this diverse information into a coherent, comprehensive, and conflict-free explanation that provides consistent understanding across all clients.

    \item \textbf{Utilization of Uncertainty Information in FL Training:} Leveraging uncertainty information within explanations to enhance FL model training involves identifying more reliable explanations from less reliable ones. Using this information to weigh the contributions from different clients and to prioritize the more reliable ones during model aggregation is challenging.

\end{enumerate}

To address these challenges, we propose the \underline{Uncertain}ty-aware e\underline{X}plainable \underline{F}ederated \underline{L}earning (\methodname{}) method. Under \methodname{}, FL clients perform a two-step explainable method to generate explanations while making predictions. Initially, clients train deep models such as ResNet~\cite{he2016deep} to extract features from images. Then, a concept-based network~\cite{barbiero2022entropy}, capable of generating logical rules as explanations, is adopted to make predictions from these features. During FL training, clients upload their model updates along with the corresponding rules to the server. The server aggregates the models and logical rules to ensure that the global rule is complete and conflict-free. In this way, \methodname{} generates global explanations without requiring access to clients' local data. In addition, \methodname{} integrates uncertainty values calculated from the input data into each explanation. The uncertainty is a critical component that contributes to the effectiveness of a logical rule, in addition to its accuracy. During the aggregation of local logical rules, clients are weighted based on the overall performance of their rules, including uncertainty scores and accuracy. \methodname{} ensures that clients with more reliable rules have greater influence on the global rule. 

We conducted extensive experiments to evaluate the effectiveness of \methodname{} on various datasets.\footnote{\url{https://anonymous.4open.science/r/Uncertain-XFL/}} The results show that \methodname{} achieves strong model accuracy and explanation accuracy, surpassing the current state-of-the-art model that does not consider uncertainty by 2.71\% and 1.77\%, respectively. The model incorporates uncertainty into the FL aggregation process, allowing the uncertainty level of the generated explanations to guide the aggregation of FL clients. By integrating and quantifying uncertainty present in the data into the explanation process, our approach explicitly shows the explanation together with the uncertainty of the explanation, thereby enhancing XFL robustness and reliability.

\section{Related Work}


FL allows users to collaboratively train models while keeping their data private. Unlike traditional methods, sensitive information is not uploaded to a central server. Instead, users send only their model updates to a central server for aggregation. This approach helps build a global model without exposing individual data. However, this decentralized method introduces challenges, particularly in developing consistent explanation models. The diverse data distributions across clients can lead to variability in model performance and behaviors, making it difficult to create a unified explanation framework that accurately reflects decision-making processes across all clients.

To address the challenges of explainability in FL, researchers have developed mechanisms for both local and global explanations. Some studies, like \cite{fiosina2021explainable}, focus on applying XAI techniques solely to local client models, bypassing the need for global explanations. Conversely, other research \cite{fiosina2021interpretable,zhang2024lr,yang2024explainable} aims to create global explanations by aggregating individual client explanations. 

Intrinsically explainable models are utilized in these efforts. For instance, \cite{yang2024explainable} describes the use of linear models as decision-making tools. Here, clients employ fuzzy rules to adjust coefficients for the linear models based on specific conditions. During aggregation, the server examines the rules for any overlapping attributes and computes the aggregated rules using a weighted average of the original rule coefficients. In addition, post-hoc explanations play a significant role in explainable FL. In \cite{fiosina2021interpretable}, the authors use Shapley values~\cite{lundberg2017unified} to determine the importance of features in explanations. They calculate global feature importance by aggregating individual Shapley values from clients, exploiting the additive properties of these values. Moreover, concept-based models, as discussed in \cite{barbiero2022entropy}, enable the generation of rule-based explanations in FL. In \cite{zhang2024lr}, clients derive logical rules from these models and send them to the FL server, which integrates these rules using suitable logical connectors to ensure a cohesive and comprehensive global explanation.

Nevertheless, existing XFL approaches do not take into account the uncertainty information. The proposed \methodname{} method aims to bridge this important gap in the current literature.

\section{Preliminaries} \label{sec:pre}


Uncertainty in AI, referring to the degree of confidence or ambiguity that AI systems exhibit in their predictions. Providing uncertainty estimates can serve as an additional form of transparency and explanation~\cite{seoni2023application}. Recognizing and quantifying this uncertainty is crucial as it enhances the interpretability of AI systems, supports more informed decision-making, and improves overall system robustness. This is especially vital in high-stakes environments such as healthcare, autonomous driving, and remote sensing~\cite{russwurm2020model}, where decisions based on uncertain predictions can have significant consequences.
The sources of uncertainty include incomplete or noisy training data, limited domain knowledge, the inherent randomness of the model architecture, and the inherent variability of the AI environment. Depending on the source, there are two main types of uncertainty in AI.

Aleatoric uncertainty (a.k.a., statistical uncertainty) emerges from inherent noise, incompleteness, conflicts, or variability in the data. It represents uncertainty that cannot be reduced even if more data is available. For instance, in medical imaging, the quality of the image can vary due to different imaging conditions and patient movements which introduce noise into the data.

Epistemic uncertainty (a.k.a., model uncertainty) arises from insufficient knowledge within the model, poor representation of training data, or flaws in the model itself. It can be mitigated as the model acquires more information about the environment through additional data or enhanced learning algorithms. This uncertainty leads to doubts about model behavior or performance in new or unseen situations~\cite{gawlikowski2023survey}. A typical example is a model trained on data from one geographic region being used in another. The lack of knowledge about the new region introduces epistemic uncertainty.

In \methodname{}, we adopt a concept-based model inspired by \citeauthor{barbiero2022entropy} to derive logical rules from neural networks. Initially, the model extracts an $F \times C$ matrix from model parameters, where there are $F$ features and $C$ classes. It indicates the contribution of each feature $F$ to each class $C$. For each data point predicted to belong to class $c$, the model examines the corresponding row in the matrix. A feature $f_i$ is considered important for class $c$ prediction if it surpasses a threshold value $t$. Depending on whether the actual feature value exceeds the threshold, it is included in the rule as $f_i$ or $\neg f_i$. The rule for an individual data point is formed by connecting these important features using the `AND' logical connective. Subsequently, the rule for a class is constructed by combining the rules from all data points belonging to that class using the `OR' logical connective. This allows the generation of specific and explainable rules based on the significance of each feature for each class.

We exclude the $\neg$ logical connective in rules within \methodname{} for two main reasons. Firstly, from a technical perspective, most datasets, such as the CUB dataset~\cite{WahCUB_200_2011}, provide comprehensive labelling of features across all potential categories within a feature genre. For example, the sizes range from ``very small (3 - 5 in)" to ``very large (32 - 72 in)", covering nearly all possible size variations. This extensive categorization makes the use of negative forms of features unnecessary. Secondly, from a psychological perspective, people generally prefer defining rules using the positive form of a feature because it is easier to understand. Describing a bird as ``medium (9 - 16 in)" is more intuitive than indicating ``NOT very small (3 - 5 in)".
Furthermore, while studies like \cite{barbiero2022entropy} and \cite{zhang2024lr} include negative forms of features in their rules based on concept-based model analysis, we contend that the importance attributed to a feature's absence might actually be influenced by the presence of a related feature. For example, the significance of a bird not being ``very small (3 - 5 in)" could actually reflect the predominance of the feature ``large (16 - 32 in)".

\section{Explainable FL with Uncertainty}

In this section, we introduce \methodname{}, a first-of-its-kind XFL framework that incorporates the uncertainty of data. It considers uncertainty as a measure for assessing the reliability of explanations and effectively handles conflicts during the aggregation of explanations.

\subsection{Overview of \methodname{}}

\begin{algorithm}[!t]
\caption{\methodname{}}
\label{alg:algorithm}
\textbf{Input}: $K$ clients, each holding a set of local data; a server, holding a set of data for validation and testing\\
\textbf{Output}: Global rules for the server; local models and rules for clients
\begin{algorithmic}[1] 
\WHILE{Global model has not achieved the target performance on the validation set \AND max training rounds have not been reached}
    \STATE \textbf{For each FL client $k, k\in\{1, \dots ,K\}$}:
        \STATE Trains the local model;            
        \STATE Generates logic rules $r_{k}^{c}$ for $c\in\{1,\dots, C\}$ classes and calculate the uncertainty $u_{k}^{c}$  for logic rules;
        \STATE Uploads the local model and rules with uncertainty information to the FL server;
    \STATE \textbf{FL Server}:
        \STATE Rank and select the received client rules based on the rule uncertainty and rule accuracy;
        \STATE Aggregates selected clients' local rules;
        \STATE Calculates client weights $\{w_{1}, \dots, w_{K}\}$ based on the times their rules being aggregated in the global rule;
        \STATE Aggregates the local models based on the assigned weights;
        \STATE Sends the global model back to the clients;
    \STATE \textbf{For each FL client $k$}: Receives the global model and continues training for the next round;
\ENDWHILE
\end{algorithmic}
\end{algorithm}

\begin{figure}[t]
    \centering
    \includegraphics[width=1\linewidth]{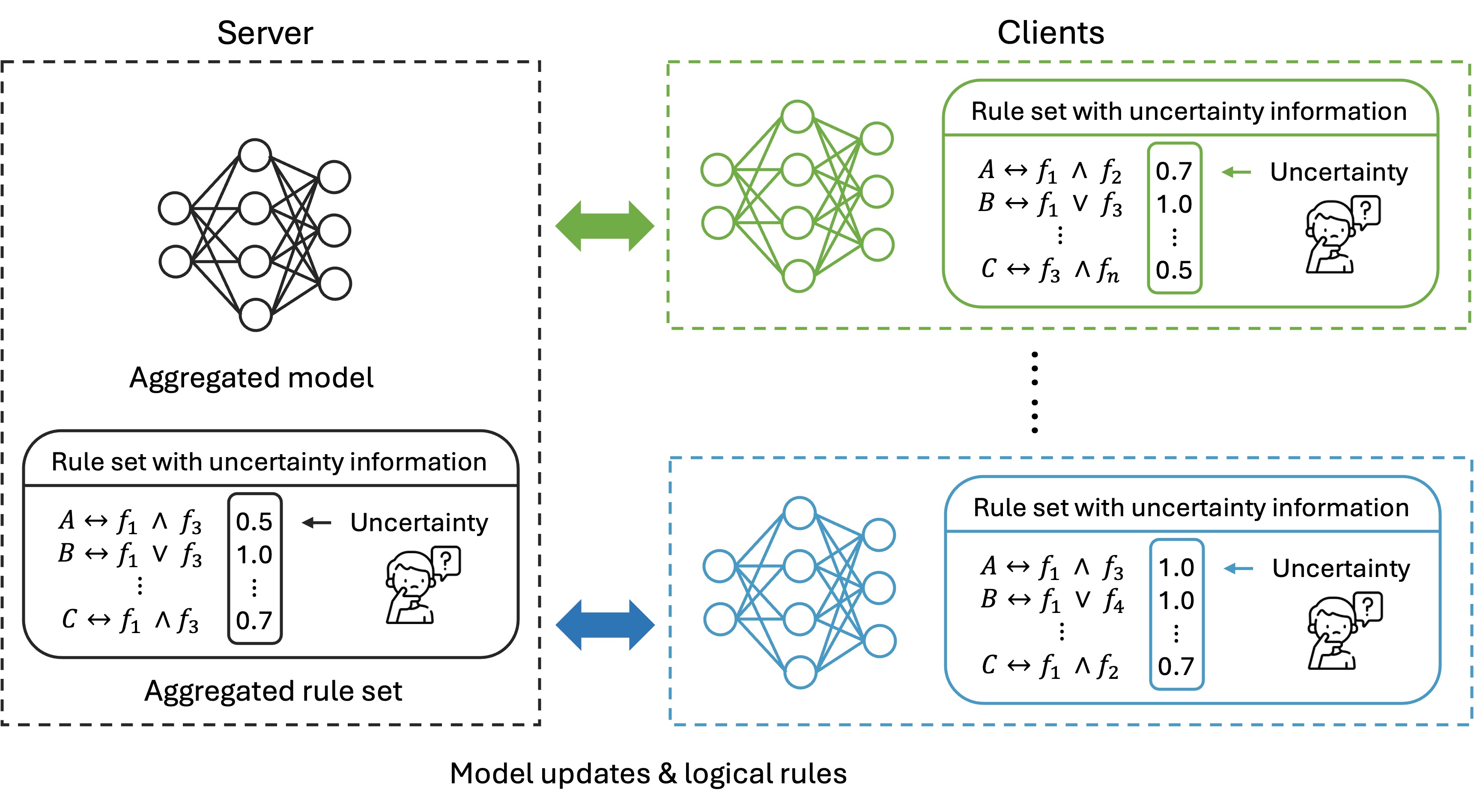}
    \caption{The overall structure of \methodname{}.}
    \label{fig:overview}
\end{figure}

Figure~\ref{fig:overview} illustrates the structure of \methodname{}. Unlike traditional FL frameworks, both the server and clients in \methodname{} maintain an explanation set in addition to the FL models. This explanation set comprises logical rules extracted from the models, which illustrates the decision-making process. As described in Algorithm~\ref{alg:algorithm}, during training, FL clients upload both the explanation rule set and the model updates to the FL server. The server then aggregates the model updates and the rule sets in a manner that avoids conflicts. Furthermore, the uncertainty level associated with each rule set is provided alongside the rules themselves, offering a transparent method for the server and stakeholders in the FL system to assess the reliability of the rules.

\subsection{Calculation of Uncertainty in Logical Rules}\label{subsec:uncertainty-cal}

\begin{figure*}[!t]
    \centering
    \includegraphics[width=0.9\linewidth]{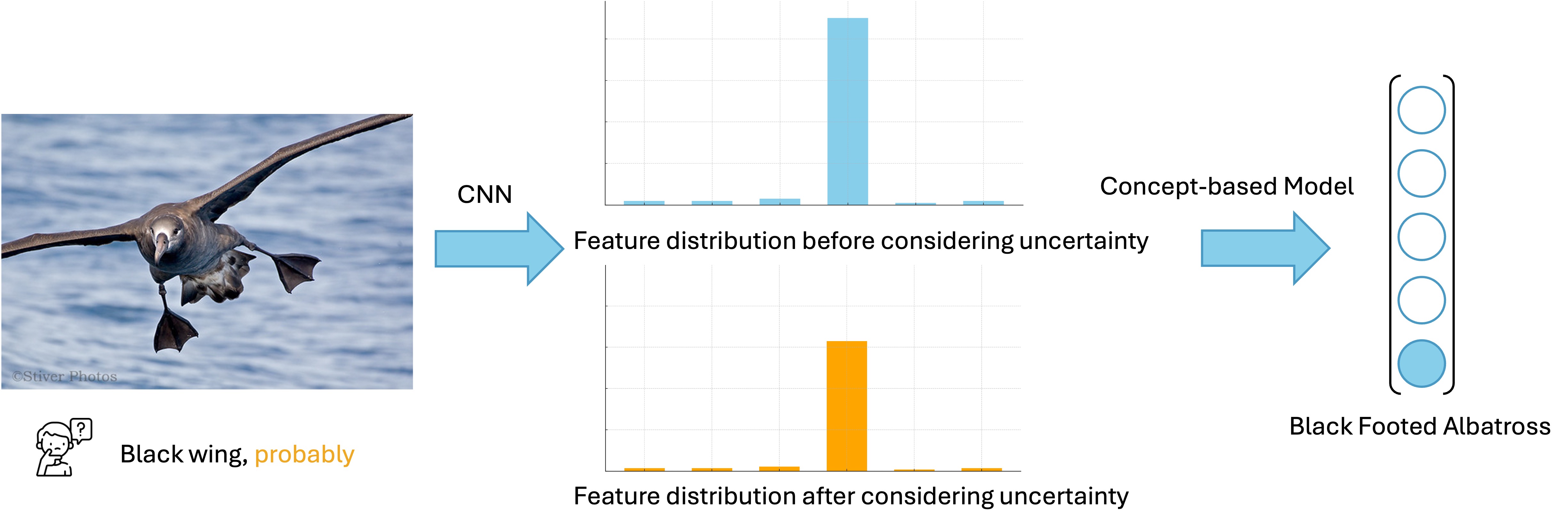}
    \caption{The workflow of client models in \methodname{}.}
    \label{fig:workflow}
\end{figure*}

In existing works~\cite{barbiero2022entropy,zhang2024lr}, the ground truth for feature values is binary (0 or 1), indicating simply whether a feature exists in an image without considering the labeller's confidence. As shown in Figure~\ref{fig:workflow}, \methodname{} addresses this limitation by incorporating uncertainty based on the labeller's confidence in identifying features. This uncertainty arises when a labeller is not entirely sure about the existence of a feature in an image. To quantitatively express this uncertainty, we assign discrete values to the ground truth. For instance, if the labeller is ``somewhat sure" that a feature exists, the feature value is set to 0.7 to represent this level of uncertainty.

To incorporate the uncertainty information into the feature values, we modify the predicted feature vector for each data point $d_i$. Assuming the predicted feature vector for data point $d_i$ is $v_i$, $v_i$ is a $F$-dimensional vector of values between 0 and 1 where $F$ is the total number of features. We then introduce an uncertainty vector $u_i$ of the same length as $v_i$. The feature vector augmented with uncertainty information, $\hat{v}_i$, is then calculated as:
\begin{equation}
\hat{v}_i = v_i \odot u_i
\end{equation}
where $\odot$ denotes element-wise multiplication.

In previous works, $v_i$ is directly sent to the concept-based network for predicting the final classes. In \methodname{}, we input $\hat{v}_i$ instead, allowing the uncertainty introduced by the labeller during the labeling of features to be conveyed through to the prediction.

If data point $d_i$ is predicted to belong to class $c$, and feature $f_j$ is activated during this prediction, the uncertainty value of feature $f_j$ is $\hat{v}_{i}^{j}$, the $j$-th value in the vector $\hat{v}_i$. If the rule $r_i$ for data point $d_i$, class $c$ contains multiple activated features, the final uncertainty score $U_{r}^{c}$ of the sample-level rule $r_i$ for class $c$ is calculated as the geometric mean of the uncertainty values of the activated features:
\begin{equation}
U_{r}^{c} = (\prod_{j=1}^m \hat{v}_{i}^{j})^{1/m}
\end{equation}
where $m$ is the number of activated features in rule $r_i$.

\subsection{Handling Conflicts in Explanation Aggregation}

Explanation aggregation in \methodname{} takes place at two levels: 1) within a client to combine data-point-level rules, and 2) across clients to establish a global logical rule. Conflicts might emerge during the aggregation process. For example, one rule could state ``Black Footed Albatross $\leftrightarrow$ Wing Color Black'', while another might contradict it with ``Black Footed Albatross $\leftrightarrow$ Wing Color Gray''. To resolve these conflicts and achieve coherent aggregation, we explore four potential aggregated outcomes: ``Black Footed Albatross $\leftrightarrow$ Wing Color Black'', ``Black Footed Albatross $\leftrightarrow$ Wing Color Gray'', ``Black Footed Albatross $\leftrightarrow$ Wing Color Black $\wedge$ Wing Color Gray'' and ``Black Footed Albatross $\leftrightarrow$ Wing Color Black $\vee$ Wing Color Gray''. We aim to select the most effective explanation, taking into account both the conflicts and the performance of the rules, which includes evaluating rule accuracy on the validation dataset and the rule uncertainty level.

To manage conflicts effectively, we first identify the root causes of feature-level conflicts. When labeling the features of a given image, features can be classified into two types: independent features, where the presence of one does not influence the presence of another, and correlated features, which imply that if one feature appears, some other features are likely not to appear. For instance, in the CUB dataset, features are organized into groups like wing colors, wing shapes, and other characteristics. Specifically, within the foot color category, it is uncommon for two different foot colors to appear in the same image, nor should they be connected by $\wedge$ in the rules. Thus, for the conflicts mentioned earlier, the combination of the two conflicting rules should not employ the $\wedge$ operator.

When features in the same group appear in the rules to be aggregated, we avoid using $\wedge$ to manage the conflict. Subsequently, we decide whether to combine the rules using $\vee$ or to retain one of the rules as the aggregated rule. Assume $acc_1$, $acc_2$, and $acc_{1\vee2}$ represent the accuracy values of the two rules separately and combined using $\vee$, tested on the validation dataset. Similarly, $u_1$, $u_2$, and $u_{1\vee2}$ represent the uncertainty scores of the rules, with $u_{1\vee2}$ calculated as the mean of $u_1$ and $u_2$. The rules are ranked based on the product of rule accuracy and uncertainty. To form a new rule, the original rules are sequentially added while checking if there is any improvement in accuracy on the validation set.

\subsection{Uncertainty-aware Rule and Model Aggregation}

In previous research on logical rule-based XFL~\cite{zhang2024lr}, the server used the beam search algorithm~\cite{lowerre1976harpy} to identify the best combination of rules. Though it saves time compared to testing every possible rule combination, it incurs memory costs by training the beam-sized best performing rule set in every step. To address this limitation, we introduce a greedy uncertainty-guided aggregation method in \methodname{}.

The selection of rule sets for aggregation is determined by rule accuracy and rule uncertainty. After a training round, clients send their model updates along with rule information back to the server. The rule information includes the rules for different classes, the accuracy of these rules tested on the clients' local test datasets, and the uncertainty score of each rule. For each client $k$, the rule corresponding to class $c$ is denoted as $r_{k}^{c}$. The accuracy of rule $r_{k}^{c}$, tested on the local dataset of client $k$, is denoted as $acc_{k}^{c}$. The uncertainty value of the rule is $u_{k}^{c}$.
The server groups clients that have submitted the same rules for class $c$. It then ranks these rules based on:
\begin{equation}
R_k = \frac{\sum_{k=1}^{n} acc_{k}^{c} \cdot u_{k}^{c}}{n}, 
\end{equation}
where $n$ is the number of clients contributing to the rule $r_{k}^{c}$. The server selects the top $m$ rules for each class based on this ranking.

When aggregating the top $m$ rules, the process begins with the highest-ranked rule and continues sequentially. If subsequent rules overlap in feature groups with previously aggregated rules (i.e., indicating potential feature conflicts), the server employs the `OR' logical connective to mitigate these conflicts. Conversely, if there are no overlapping feature groups (i.e., no conflict), the `AND' logical connective is used to combine these rules as they are likely to complement each other. In addition, a validation dataset is maintained by the server to assess the performance of each new rule. A rule is only integrated into the global rule set if its inclusion improves model performance on the validation set.

In addition, $t_k$ records the number of times a client $k$'s rules are ranked within the top $m$ for global rule aggregation during a training iteration. The weight assigned to $k$ for model aggregation in this training round is calculated as:
\begin{equation}
w_{k} = \frac{t_{k}}{\sum_{i=1}^{K} t_{i}}.
\end{equation}
This weight reflects the frequency for which a client's rules are regarded as important. In this way, it ensures that more reliable contributors exert greater impact on the FL model.

\section{Experimental Evaluation}

\subsection{Dataset Description}

\begin{figure}[t]
    \centering
    \begin{subfigure}{0.32\linewidth}
        \centering
        \includegraphics[width=\linewidth]{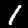}
        \caption{``Definitely'' (100\% confidence)}
    \end{subfigure}%
    \hfill 
    \begin{subfigure}{0.32\linewidth}
        \centering
        \includegraphics[width=\linewidth]{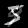}
        \caption{``Probably'' (70\% confidence)}
    \end{subfigure}%
    \hfill
    \begin{subfigure}{0.32\linewidth}
        \centering
        \includegraphics[width=\linewidth]{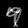}
        \caption{``Guessing'' (50\% confidence)}
    \end{subfigure}
    \caption{MNIST images with different level of uncertainty.}
    \label{fig:three_figs}
\end{figure}

Following the dataset settings in works~\cite{barbiero2022entropy,zhang2024lr}, we evaluate \methodname{} on the CUB~\cite{WahCUB_200_2011} and MNIST(Even/Odd)~\cite{lecun1998mnist} datasets. These datasets adhere to the ``image $\rightarrow$ features $\rightarrow$ classes" structure. Specifically, in the MNIST(Even/Odd) dataset, the features are the digits in the pictures, and the classes are determined by whether the digit is even or odd. In CUB, the features include various bird characteristics, with classes being specific bird categories. 

In the CUB dataset, uncertainty is explicitly introduced by the labeller, who assigns uncertainty scores to feature groups. The levels of uncertainty include ``definitely", ``probably", ``guessing", and ``not visible". Features within the same group share identical uncertainty scores. For example, a labeller might mark ``probably" for all features under the ``bill color" feature group for a bird image.

The original MNIST dataset does not contain uncertainty information. To introduce uncertainty, we estimate the uncertainty in MNIST(Even/Odd) as illustrated in Figure~\ref{fig:three_figs}.  We introduce uncertainty into the MNIST(Even/Odd) dataset by overlaying original images with images of other digits. Each image in the MNIST dataset was assigned a 50\% probability of remaining unchanged and a 50\% probability of being combined with images from other digit classes. The unchanged images are regarded as having an uncertainty level of ``definitely". Regarding the proportion of the overlay, we maintained half of the images at 70\% original and 30\% other digits, and the other half at 50\% original and 50\% other digits. These are regarded as uncertainty levels ``probably" and ``guessing" separately.

For both datasets, we established a federated data setting~\cite{mcmahan2017communication} with a uniform distribution of data across FL clients. We randomly divided the dataset evenly among all clients. In our experiment, 10 clients are involved in FL training.

\subsection{Comparison Approaches}

As this is the first work considering uncertainty information in explainable FL settings, there are no previous works to compare. Thus, we compare \methodname{} with a previous explainable FL framework LR-XFL~\cite{zhang2024lr} that does not consider uncertainty in the explanation. We remove the $\neg$ form in rules generated in LR-XFL due to the reasons mentioned in the Preliminaries section. We also conducted two additional experiments to demonstrate the effectiveness of the uncertainty information in \methodname{}. 

Firstly, we use FedAvg~\cite{mcmahan2017communication} instead of uncertainty-weighted aggregation during the federated aggregation step. FedAvg is a widely used method to aggregate model updates from FL clients. In FedAvg, clients are assigned the same importance during the aggregation, regardless of their contribution to the global model. The second experiment is to completely remove the uncertainty information and use FedAvg for aggregation. In \methodname{}, the uncertainty information is added to the system by multiplying the predicted feature with the human-labelled uncertainty level. For example, in the CUB dataset, there are four levels of uncertainty when the labeller labels the features in the images. We map ``definitely" to 1, ``probably" to 0.7, ``guessing" to 0.5, and ``not visible" to 0. In the MNIST(Even/Odd) dataset, the uncertainty is the percentage the digits are being stacked. When completely removing the uncertainty information, the feature either exists or does not exist in the images, without any uncertainty value.

\subsection{Evaluation Metrics}

We evaluated \methodname{} and baseline models using classification accuracy, rule accuracy, rule fidelity and rule uncertainty. Classification accuracy assesses the model's prediction accuracy. Rule accuracy, rule fidelity and rule uncertainty evaluate the effectiveness of explanations.

\begin{enumerate}

    \item \textbf{Classification Accuracy}: it assesses the consistency between the predictions made by the model and the ground truth classes. It is calculated by dividing the total number of correct predictions with the total number of predictions made.

    \item \textbf{Rule Accuracy}: it measures how consistently the predictions of the rules align with the actual ground truth. For instance, if a data point belongs to class $c$, and its rule predicts it to be class $c$, it positively contributes to the rule accuracy for class $c$. Similarly, if the data point does not belong to class $c$, and its rule predicts it not to be class $c$, it also positively contributes to the rule accuracy for class $c$. Assuming there are $M$ data points belonging to class $c$, of which $m$ are correctly predicted as class $c$ by the rules, and there are $N$ data points not belonging to class $c$, of which $n$ are predicted as not class $c$, the rule accuracy for class $c$ can be calculated as:
    \begin{equation}
        RAcc_c = \frac{m+n}{M+N}.
    \end{equation}
    The overall rule accuracy for the model is then determined by averaging the $RAcc_c$ values for all $C$ classes.

    \item \textbf{Rule Fidelity}: it assesses the consistency between rule predictions and model predictions. Unlike rule accuracy, which compares rule predictions to the ground truth, rule fidelity compares them with the predictions by the model. 


    \item \textbf{Rule Uncertainty}: it evaluates the uncertainty level of the global aggregated rules. It is calculated as the average uncertainty across all $C$ global rules, corresponding to the $C$ different classes.
\end{enumerate}

The evaluation metrics are calculated using a test dataset stored on the server, which makes up 5\% of the total data. This test dataset is distinct from the training data, ensuring independent validation of model performance.

\begin{table*}[t]
\centering
\caption{Experiment results. The best performance is marked in bold. `-' means the given evaluation metric is not applicable.} 
\begin{tabular}{|cl|c|c|c|c|}
\hline
 &     & \methodname{}   & \methodname{}-FedAvg  & \methodname{} w/o & LR-XFL \\
  &     &    &   & Uncertainty & \\ \hline

\multicolumn{1}{|c|}
{\multirow{4}{*}{CUB}} 
 & model \mbox{accuracy} & \textbf{90.34} \% & 87.63\%  & 88.98\% & 87.96\%\\ 
\multicolumn{1}{|c|}{}
 & rule \mbox{accuracy}  & \textbf{90.84} \% & 89.16\% & 87.85\% & 89.26\%\\ 
\multicolumn{1}{|c|}{}
 & rule \mbox{fidelity} & 99.56\% & 99.54\% & 99.51\% & \textbf{99.63\%} \\ 
\multicolumn{1}{|c|}{}
 & rule \mbox{uncertainty} & 73.44\% & \textbf{73.98\%} & - & - \\ 
\hline

\multicolumn{1}{|c|}
{\multirow{4}{*}{MNIST}} 
 & model \mbox{accuracy} & 95.71\% & 91.71\% & 97.40\% & \textbf{97.54\%} \\ 
\multicolumn{1}{|c|}{}
 & rule \mbox{accuracy} & 91.08\% & 91.27\% & 93.26\% & \textbf{95.92\%} \\ 
\multicolumn{1}{|c|}{}
 & rule \mbox{fidelity} & 91.40\% & 91.42\% & 95.23\% & \textbf{98.01\%} \\ 
\multicolumn{1}{|c|}{}
 & rule \mbox{uncertainty} & \textbf{97.19\%} & 95.06\% & - & - \\ 
\hline

\end{tabular}%
\label{result1}

\end{table*}

\subsection{Results and Discussion}
Table~\ref{result1} shows the comparison results between \methodname{} and the baselines. 

\subsubsection{Model Performance}

\methodname{} achieved the highest accuracy on the CUB dataset, reaching 90.34\%. This performance surpasses \methodname{}-FedAvg by 3.09\%, demonstrating the effectiveness of utilizing performance metrics from client-generated explanations, including accuracy and uncertainty, to guide the global model towards improved training outcomes. In addition, \methodname{} outperformed \methodname{} w/o Uncertainty and LR-XFL by 1.53\% and 2.71\% respectively, highlighting the benefits of incorporating uncertainty information from labellers during data labeling, which provides valuable insights for training more precise models.

In the MNIST dataset, \methodname{} also shows strong performance with a 95.71\% accuracy rate, although it slightly trails behind models that do not consider uncertainty information, such as LR-XFL and \methodname{} w/o Uncertainty. This discrepancy might be attributed to the fact that the uncertainty in MNIST was artificially simulated. In practical scenarios, even if an MNIST image is overlaid with another digit image in a 70\% to 30\% ratio, humans can still make definitive judgements about the image. However, by introducing uncertainty information into the predicted feature distribution, we inadvertently lower model confidence in making correct decisions, thus potentially reducing overall model accuracy.

\subsubsection{Rule Performance}

\begin{table*}[t]
    \centering
    \caption{Comparison of rules generated by \methodname{} and LR-XFL for identifying the Common Yellowthroat.} \label{tab:rule}
    \begin{tabular}{|l|l|}
        \hline
          & Rule for Common Yellowthroat \\ \hline
          \methodname{} & $ \text{throat\_color-yellow} \wedge \text{forehead\_color-black} \wedge  \text{primary\_color-yellow} \wedge  \text{wing\_pattern-solid}$ \\ \hline
          LR-XFL & $ \text{forehead\_color-black} \wedge \text{under\_tail\_color-yellow} \wedge \neg \text{crown\_color-black }$ \\ \hline
    \end{tabular}
\end{table*}

On the CUB dataset, \methodname{} achieves the highest rule accuracy at 90.84\%, outperforming other models, including LR-XFL, the second-best model, by 1.77\%. This demonstrates the accuracy of its logic-based explanations in alignment with ground truth classifications. For rule fidelity, \methodname{} reaches 99.56\%, closely following the top performer, LR-XFL, by a slight margin of 0.07\%. This high fidelity indicates that the predictions from \methodname{}'s rules are highly consistent with its model predictions, confirming the reliability of its logical rules. Additionally, superior performance compared to UncertainXFL-FedAvg and No-uncertainty XFL highlights the benefits of incorporating uncertainty information and utilizing uncertainty-weighted rule aggregation.

On the MNIST dataset, the rule performance of \methodname{} does not achieve the levels observed in models that do not incorporate uncertainty information. We infer that this discrepancy is partly due to the simulated uncertainty, which also adversely affects the model's accuracy.

The rule uncertainty metric is not used to assess the quality of the rules directly but to provide an explicit evaluation of the uncertainty levels inherent in the global rules. This metric is derived from the uncertainty present in the component features and the individual rules aggregated from various clients. For both datasets, the rule uncertainty values for \methodname{} and UncertainXFL-FedAvg are closely aligned. This similarity may stem from the fact that, although they formulate rules slightly differently, both approaches rely on comparable sets of features to generate these rules, leading to similar uncertainty scores.

\begin{figure}[!t]
    \centering
    \includegraphics[width=0.6\linewidth]{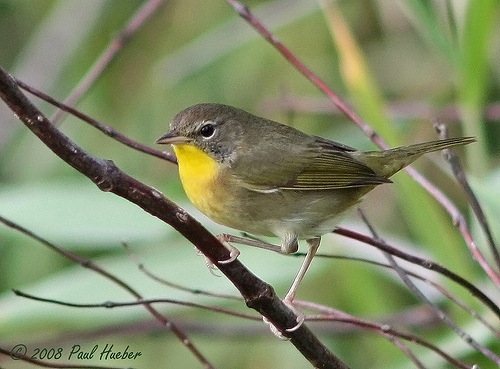}
    \caption{An image of Common Yellowthroat}
    \label{fig:bird}
\end{figure}

\subsection{Rule Comparison without``Not''}

In ~\cite{barbiero2022entropy,zhang2024lr}, $\neg$ is utilized in the rules. However, as discussed in the Preliminaries section, we chose to exclude the use of the $\neg$ form in rules within \methodname{}. This decision is based on the premise that avoiding the $\neg$ form presents a more intuitive and understandable explanation for users.

To illustrate that omitting the $\neg$ form provides clearer explanations, we provide example rules extracted from \methodname{} and LR-XFL~\cite{zhang2024lr} for the bird species, Common Yellowthroat, as illustrated in Table~\ref{tab:rule}. 
It is evident that including $\neg \text{crown\_color-black}$ in a rule provides a less effective description of the bird's features. As depicted in Figure~\ref{fig:bird}, the rule from \methodname{} captures the key features of the Common Yellowthroat more accurately. Furthermore, eliminating the use of $\neg$ in components of our rules simplifies comprehension and offers a more logical and direct description, making it easier for humans to understand and interpret the features.

\section{Conclusions and Future Work}

In this paper, we proposed \methodname{}, a first-of-its-kind XFL method that takes uncertainty into account. Under \methodname{}, explanations are provided in the form of logical rules, making them easy for individuals to interpret. These explanations exist both locally at the FL client side and globally on the FL server side. The global explanation ensures the aggregation of local rules in a complete and conflict-free manner, offering users an overall understanding of how the model makes decisions without accessing local private data. In addition, the uncertainty information for the generated explanations is provided, which is beneficial for stakeholders to gain insight into the confidence of the generated explanations to make informed decisions. The uncertainty information also guides the FL training to enable clients which are more confident about their decisions make a big impact on model performance.

In the future, we plan to incorporate model uncertainty alongside the aleatoric uncertainty currently used. Model uncertainty stems from the model's limitations (e.g., inadequate knowledge, sub-optimal data representation), and can be reduced through additional data or improved algorithms. Our aim is to create a more robust FL system that accounts for both types of uncertainty, providing more reliable and precise explanations across different scenarios. This enhancement will involve integrating advanced methods to measure and include model uncertainty in the global FL model, thereby enhancing system effectiveness and dependability, especially when applied to new environments.




\bibliographystyle{named}
\bibliography{ijcai25}

\end{document}